%% file: main.tex
\documentclass{article}

\PassOptionsToPackage{numbers,sort&compress}{natbib}
\usepackage[main, final]{neurips_2026}

\usepackage[utf8]{inputenc}
\usepackage[T1]{fontenc}
\usepackage{hyperref}
\usepackage{url}
\usepackage{booktabs}
\usepackage{amsfonts}
\usepackage{amsmath,amssymb,amsthm,mathtools,bm}
\usepackage{nicefrac}
\usepackage{microtype}
\usepackage{xcolor}
\usepackage{enumitem}
\usepackage{graphicx}
\usepackage{subcaption}

\hypersetup{colorlinks=true, linkcolor=blue!50!black, citecolor=blue!50!black, urlcolor=blue!50!black}
\setcitestyle{numbers,square,sort&compress}

\newtheorem{theorem}{Theorem}[section]
\newtheorem{proposition}[theorem]{Proposition}

\theoremstyle{remark}
\newtheorem{remark}[theorem]{Remark}

\newcommand{\E}{\mathbb{E}}
\newcommand{\Risk}{\mathcal{R}}
\newcommand{\calD}{\mathcal{D}}
\newcommand{\calM}{\mathcal{M}}

\newcommand{\calN}{\mathcal{N}}

\newcommand{\doop}{\mathrm{do}}
\DeclareMathOperator*{\argmax}{arg\,max}

\title{Specialist Mediators:\\
Causal Localization of Dense Fine-Tuning}

\author{%
  Amir Asiaee \\
  Department of Biostatistics \\
  Vanderbilt University Medical Center \\
  Nashville, TN 37203, USA \\
  \texttt{amir.asiaeetaheri@vumc.org}
}

\begin{document}

\maketitle

\begin{abstract}
Dense fine-tuning can turn a base language model into a domain specialist, but task accuracy alone does not reveal which internal computations mediate the new behavior. We introduce \emph{specialist mediators}, the internal components through which a dense specialist's lift over its base causally flows. We identify these mediators by patching specialist activations into the base model on the same input and asking which patches recover the specialist's in-domain score while minimally perturbing off-domain behavior. This separates where the specialist behavior is localized, how much of it can be recovered by activation patching, and whether a sparse trainable update can implement the same behavior, three questions that are often conflated. We formalize this with an exact recovery result, a first-order approximation for joint patching, and a conditional bound showing when patching recovery can transfer to localized training. In synthetic transformers and structural causal models with planted mediators, patching recovers most planted mediators even when activation-magnitude baselines are misled by low-output distractors. In Pythia-410M specialists for math (GSM8K), multitask knowledge (MMLU), and biomedical QA (PubMedQA), single-patch localization ranks first on every domain and seed by area under the recovery curve. Five components recover more than 90\% of the dense specialist's lift in answer-sequence likelihood on MMLU and PubMedQA, and a Pythia-1.4B PubMedQA specialist shows the same concentration. A comparison of LoRA adapters at exactly matched parameter counts then separates localization from trainability, showing that a few components can carry the specialist's score while adapter performance depends on more than where the adapter is placed.
\end{abstract}

\section{Introduction}

Fine-tuning can turn a general language model into a domain specialist, but task accuracy does not reveal which internal computations now carry the domain-specific lift. Did fine-tuning change a few routing components, redistribute the computation broadly, or mainly alter output-side calibration? This distinction matters for interpretability, auditing, and sparse adaptation, because deciding where to edit, monitor, or retrain a specialist requires a causal map of where its extra behavior flows.

We study this as a \emph{model-difference localization} problem. By localization, we mean identifying a small set of internal components, such as attention heads or MLP blocks, that explain a target behavior. We call heads and MLP blocks \emph{components} and reserve \emph{mechanism} for a computational subgraph. Here the target is the \emph{difference} between two trained models, a base $f_0$ and a dense specialist $f_e$. We ask which components causally mediate the performance gap $f_e{-}f_0$.

Our basic intervention is \emph{activation patching}, which runs two forward passes on the same input and replaces selected internal activations in one model with the corresponding activations from another. We use cross-model activation patching (CMAP) \citep{prakash2024finetuning}, in which both passes share an input and the activations come from a fine-tuned model. We patch specialist activations into the base and measure how much of the dense-specialist lift returns while off-domain behavior stays close to base. A good \emph{specialist mediator set} is one through which the in-domain lift causally flows. The definition is operational, so a mediator set is characterized by the effect of this intervention and need not be unique.

This mediator set supports three related but distinct goals. The first is \emph{localization}, finding the components through which the specialist's extra behavior flows. The second is \emph{patching recovery}, checking whether inserting only those specialist activations into the base recovers the specialist's behavior. The third is \emph{localized training}, updating only parameters near the mediator set instead of fine-tuning the whole model. The third goal requires more, since patching produces a chimera, base weights with specialist activations inserted, and a small trainable parameter class must be able to implement the same computation. Our transfer theorem makes this realizability requirement explicit, and our experiments measure all three goals and show that they can come apart.

The theory follows this sequence from localization to recovery to training. We start with the ideal case, in which fine-tuning changed the computations of only a few components and left the rest of the network as it was. The unchanged components then compute the specialist's values from the specialist's inputs, so patching the changed components reproduces the specialist on every input. Dense fine-tuning usually spreads its changes more widely, so a small patch recovers the specialist only approximately. For this case we write the recovery of a patched set as a sum over its components, where each term is the score's sensitivity to that component times the change in its activation, plus a remainder controlled by the curvature of the score. A component's sensitivity depends on what else is patched, since a patch on a downstream component overwrites what an upstream component sends through it. Scores measured one component at a time can therefore count an effect twice, so we use them to rank candidates and measure the recovery of the resulting sets directly. Training raises a different question, because a patch inserts specialist activations at run time while an adapter must produce them through its own parameters. Our transfer bound limits a localized adapter's excess risk over the dense specialist by two gaps, the distance from the patched model's outputs to the specialist's and the distance from the trainable class to the patched model. Localization speaks to the first gap, and the choice of trainable class governs the second.

The experiments follow this progression. Synthetic transformers and structural causal models test localization when the true mediators are known, and patching stays closer to the planted set when activation magnitudes are confounded by distractors with attenuated outputs. Pythia-410M specialists \citep{biderman2023pythia} test patching recovery on math word problems (GSM8K \citep{cobbe2021gsm8k}), broad multitask knowledge (MMLU \citep{hendrycks2021mmlu}), and biomedical question answering (PubMedQA \citep{jin2019pubmedqa}), with a Pythia-1.4B check on PubMedQA. The localized quantity is the teacher-forced negative log-likelihood (NLL) of the answer sequence, the reference answer followed by an end-of-sequence (EOS) token. The results therefore describe fine-tuning-induced changes in answer-sequence likelihood. A localized low-rank adaptation (LoRA) comparison then probes the training step.

\paragraph{Contributions.}
\begin{enumerate}[leftmargin=18pt, itemsep=1pt]
    \item We define \textbf{specialist mediator sets} by interventional recovery and off-domain interference (Section~\ref{sec:problem-setup}).
    \item We give theory for the localization--recovery--training sequence, with exact recovery under structural localization, a first-order expansion for joint patching, and a conditional transfer bound that separates patching error from localized-class realizability (Section~\ref{sec:theory}).
    \item We evaluate localization and patching recovery on synthetic planted mediators and three Pythia-410M LM domains. Single-patch localization ranks first on every (domain, seed) AURC, and the concentration persists at 1.4B (Sections~\ref{sec:e1}--\ref{sec:scale}).
    \item We compare localized LoRA supports at exactly matched parameter counts and show that localization and trainability are distinct questions (Section~\ref{sec:e3}).
\end{enumerate}

\section{Related work}
\label{sec:related-work}

\paragraph{Mechanistic interventions and causal mediation.}
Interchange interventions, activation patching, and path patching are standard primitives for localizing computations in neural networks \citep{geiger2021causal, meng2022locating, conmy2023acdc, goldowskydill2023path, geiger2024finding}. A related causal-mediation line treats internal components as mediators between inputs and model behavior, for example in analyses of gender bias, syntactic agreement, and arithmetic reasoning in language models \citep{vig2020investigating, finlayson2021causal, stolfo2023mechanistic}. Sparse intervention methods also search for small sets of states whose removal or replacement changes a behavior \citep{de2022sparse}. Attribution patching and AtP* approximate patch effects with gradients to scale circuit search \citep{nanda2023attribution, syed2024attribution, kramar2024atp}. Certified interventional fidelity adds anytime-valid confidence sequences for such estimates \citep{asiaee2026certified}. These tools hold the weights fixed and contrast inputs. Causal tracing restores clean activations into a run on a corrupted input \citep{meng2022locating}, path patching restricts the same swap to chosen paths \citep{goldowskydill2023path}, and attribution patching linearizes it \citep{nanda2023attribution}. Cross-model activation patching (CMAP) \citep{prakash2024finetuning} instead fixes the input and takes activations from a fine-tuned model, and it attributed a fine-tuning gain on entity tracking to a step of an existing circuit. We use the same operator (Table~\ref{tab:contrast}) and add a recovery-and-interference selection objective over all candidate components, a conditional analysis of when patching transfers to localized training, and an empirical comparison of the two.

\begin{table}[t]
\centering
\caption{\textbf{What is contrasted and what is intervened on.} Patching tools share the activation-swap primitive, and this paper uses the cross-model contrast of CMAP.}
\label{tab:contrast}
\small
\begin{tabular}{@{}p{0.25\linewidth}p{0.24\linewidth}p{0.17\linewidth}p{0.24\linewidth}@{}}
\toprule
Setting & Contrast & Intervention target & Question answered \\
\midrule
Activation and path patching, causal tracing \citep{zhang2024bestpractices, goldowskydill2023path, meng2022locating} & clean vs.\ corrupted input, same weights & activations or paths & Which sites carry an input-level effect? \\
Skill localization by grafting \citep{panigrahi2023skill} & base vs.\ fine-tuned weights & parameter values & Which parameter deltas recover a task? \\
CMAP \citep{prakash2024finetuning} & fine-tuned vs.\ base weights, same input & cross-model activations & Which circuit step explains a fine-tuning gain? \\
This paper & base vs.\ dense specialist, same input & cross-model activations & Which components carry the lift, and does adapting them help? \\
\bottomrule
\end{tabular}
\end{table}

\paragraph{Skill localization and model differences.}
The closest work in problem formulation asks where newly acquired skills live in a fine-tuned model by \emph{grafting} fine-tuned parameter values into the base and measuring recovered task performance \citep{panigrahi2023skill}. That work localizes sparse parameter deltas, whereas we localize activation-level mediators of the performance gap.

\paragraph{Localized representation tuning and PEFT.}
Adapters, LoRA, and DoRA specialize a frozen or partially frozen backbone through small trainable parameter subspaces \citep{houlsby2019adapters, hu2021lora, liu2024dora}. LoFiT is especially close in spirit, selecting task-specific attention heads and training representation offsets there \citep{yin2024lofit}. ReFT/LoReFT learns interventions on hidden representations as a parameter-efficient alternative to standard fine-tuning \citep{wu2024reft}. These methods are adaptation mechanisms. Our method is a selection and diagnostic criterion derived from a dense teacher. It first asks where the dense specialist's extra behavior is mediated, then separately asks whether a chosen localized parameter class can realize the patched computation.

\paragraph{Localize-then-correct, editing, and editability of localization.}
Causal tracing localizes where factual associations appear to be represented, and rank-one or low-rank editing methods modify weights near those locations \citep{meng2022locating, meng2023memit}. However, prior work shows that localization by causal tracing does not necessarily predict the best layer for knowledge edits \citep{hase2023localization}. Attribution-guided pruning \citep{hatefi2025attribution} uses layer-wise relevance propagation within one model to identify the parameters behind an undesired behavior and corrects the behavior by pruning that circuit. Our setting differs in both steps, since we localize a difference between two models and the downstream question is whether a trainable local class can reproduce a patched computation rather than whether deletion removes a behavior. The approaches are complementary. Our setting is broader than one factual edit, but the same lesson applies, since finding a causal site and implementing an update there are distinct claims. Theorem~\ref{thm:transfer} makes this separation explicit by isolating the patching gap from the realizability gap of the localized parameter class.

A broader survey of causal abstraction, structured pruning, parameter-efficient adaptation, and modularity appears in Appendix~\ref{app:related-broader}.

\section{Problem setup}
\label{sec:problem-setup}

\subsection{Base model, dense specialist, and patching operator}

Fix a set of environments $\mathcal{E}$ and one target environment $e \in \mathcal{E}$ with data distribution $\calD_e$ and task loss $\ell_e$. Let $f_0 := f_{\theta_0}$ be the base generalist and $f_e := f_{\theta_e}$ a \emph{dense specialist teacher}, obtained for example by full fine-tuning or a sufficiently expressive PEFT method on environment $e$. We write $\Risk_e(\theta):=\E_{(x,y)\sim\calD_e}[\ell_e(f_\theta(x),y)]$ for the risk induced by the loss.

Let $\calM$ denote the candidate components to localize over, such as attention heads, MLP blocks, channels, neurons, or learned features. We treat the forward pass as a deterministic structural causal model (SCM) whose endogenous variables are the intermediate activations and whose structural equations are the layer computations \citep{geiger2021causal, beckers2019abstracting}. The same view of a trained network as an SCM over its internal activations appears in recent work \citep{asiaee2026efficient}. For a component $m\in\calM$, write $a_m^0(x)$ and $a_m^e(x)$ for the activation produced by the base and specialist on the same input $x$.

For a set $S\subseteq \calM$, define the \emph{specialist-to-base patching operator} $P_S(x):=f_{\theta_0}^{\doop(A_S := a_S^e(x))}(x)$, that is, the base model evaluated on $x$ with the activations in $S$ replaced by the specialist's activations on that same input.

\begin{figure}[t]
  \centering
  \includegraphics[width=\linewidth]{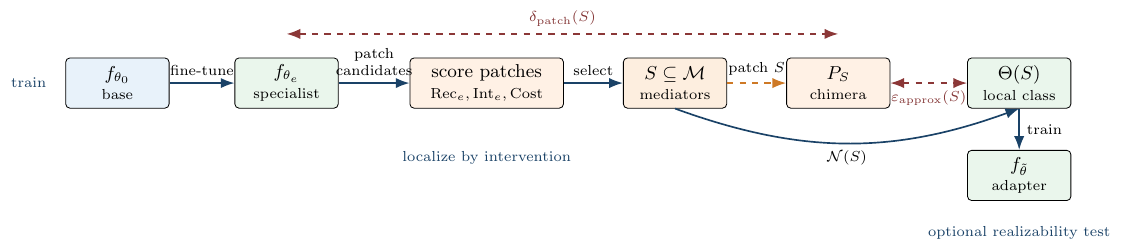}
  \caption{\textbf{Specialist localization, patching, and realizability.} We select mediators $S$ by specialist-to-base patching scores, then separate the patching gap $\delta_{\mathrm{patch}}(S)$ from the realizability gap $\varepsilon_{\mathrm{approx}}(S)$ of the localized class $\Theta(S)$.}
  \label{fig:F1_pipeline}
\end{figure}

\subsection{Recovery, interference, and mediator selection}

We use score-based quantities for selecting mediators and risk-based quantities for the transfer theorem. Let $\psi_e(f;x,y)$ be a task score where larger is better. For the theoretical results in Section~\ref{sec:theory} (which require differentiability), we take $\psi_e$ to be a differentiable surrogate such as negative loss or log-probability of the correct answer. In the language-model experiments, $\psi_e$ is the negative teacher-forced NLL of the target sequence (Section~\ref{sec:e2}). Define the dense-specialist lift $G_e:=\E_{(x,y)\sim\calD_e}[\psi_e(f_e;x,y) - \psi_e(f_0;x,y)]$, the quantity we want the mediator set to explain.

The \emph{recovery} of a patched set $S$ is $\mathrm{Rec}_e(S):=\E_{(x,y)\sim\calD_e}[\psi_e(P_S;x,y)-\psi_e(f_0;x,y)]$. There are two useful normalizations. The \emph{dense-lift recovery} is $\rho^{\mathrm{lift}}_e(S):=\mathrm{Rec}_e(S)/G_e$, defined whenever $G_e > 0$ (i.e., the dense specialist improves over the base). When $G_e \approx 0$, the specialist provides no meaningful lift and localization is not applicable. The \emph{patchable-gap recovery} is $\rho^{\mathrm{patch}}_e(S):=\mathrm{Rec}_e(S)/\mathrm{Rec}_e(\calM)$, defined when $\mathrm{Rec}_e(\calM)>0$. This second normalization is useful when the candidate set $\calM$ does not include every component that differs between base and specialist. We call $c_e := \mathrm{Rec}_e(\calM)/G_e$ the \emph{patching ceiling}, the fraction of dense lift recovered by patching all components under consideration. The term denotes this full-set reference value rather than an upper bound over subsets, since recovery need not be monotone in $S$ and either normalization can be negative or exceed one. When $\calM$ contains every component that changed, $c_e=1$ and the two normalizations coincide. In the language-model experiments, embeddings and the unembedding are left unpatched, so recovery curves are normalized by the patchable gap.

To discourage components that indiscriminately change behavior outside the specialty, define (for $|\mathcal{E}|\ge 2$) the off-domain interference $\mathrm{Int}_e(S):=|\mathcal{E}\setminus\{e\}|^{-1}\sum_{e'\neq e}\E_{(x,y)\sim\calD_{e'}}\big[|\psi_{e'}(P_S;x,y)-\psi_{e'}(f_0;x,y)|\big]$, where $\psi_{e'}$ is the score function for environment $e'$. The absolute value penalizes any off-domain behavioral change, whether helpful or harmful, since the goal is a mediator set that is specific to the target domain.

For the risk bound in Section~\ref{sec:theory}, we also track the output-space patching gap $\delta_{\mathrm{patch}}(S):=\E_{X\sim\calD_e}[\|P_S(X)-f_{\theta_e}(X)\|]$. It plays the role in the transfer bound that score recovery plays in the experiments. Outputs are a fixed representation with a specified norm, and for cross-entropy we use vocabulary logits with the Euclidean norm under the teacher-forced convention of Appendix~\ref{app:setup}.

The ideal mediator set solves
\begin{equation}
S_e^\star\in\argmax_{S\subseteq \calM}\big\{\mathrm{Rec}_e(S)-\beta_{\mathrm{int}}\,\mathrm{Int}_e(S)-\lambda\,\mathrm{Cost}(S)\big\},
\label{eq:mediator-objective}
\end{equation}
where $\mathrm{Cost}(S)$ may be cardinality, FLOPs, number of trainable parameters induced by $S$, or latency overhead.

\section{Theory}
\label{sec:theory}

This section states the steps from localization to training as three results about a set of patched components. Proposition~\ref{prop:exact} treats the ideal case, in which fine-tuning changes only a few components. Proposition~\ref{prop:linear} expands the recovery of any patched set when the changes are more widespread. Theorem~\ref{thm:transfer} moves from patching to training and bounds the risk of a localized adapter. All proofs are in Appendix~\ref{app:proofs}.

\subsection{Exact mediator sufficiency}

Suppose fine-tuning changed the computations of only the components in a set $T_e$ and left the rest of the network as it was. A patch at every component in $T_e$ gives each later computation the same inputs it sees in the specialist. Those computations are unchanged, so they produce the specialist's values all the way to the output, and the patched model reproduces the specialist exactly.

\begin{proposition}[Exact mediator sufficiency]\label{prop:exact}
Suppose the dense specialist $f_e$ is obtained from the base model $f_0$ by changing only the structural equations of a set of components $T_e\subseteq\calM$, while every other node of the computation graph, including input and output computations outside $\calM$, is unchanged. If patching is applied at the outputs of every component in $T_e$, then $P_{T_e}(x)=f_e(x)$ for every input $x$. Hence $\rho^{\mathrm{lift}}_e(T_e)=1$ whenever $G_e>0$, and $\mathrm{Int}_e(T_e)$ equals the specialist's own off-domain score change.
\end{proposition}

The result requires patching every changed computation. Our language-model candidate set $\calM$ omits the embeddings, final layer norm, and unembedding, so Section~\ref{sec:e2} reports the patching ceiling $c_e$, the recovery when every component in $\calM$ is patched.

\subsection{Linearization under joint patching}

Dense fine-tuning changes every parameter, so Proposition~\ref{prop:exact} rarely applies exactly. We then want to know how much of the lift a small set recovers and how that recovery divides among its components. A first-order expansion addresses both. Treat the score as a function of the values placed at the patched sites, with everything else recomputed. Moving those values from the base activations to the specialist's changes the score by approximately the gradient times the change, summed over the sites, and the curvature of the score controls the error. Fix $(x,y)$ and a set $S\subseteq\calM$ of distinct patch sites, write $\Delta_m(x):=a_m^e(x)-a_m^0(x)$ and $\Delta_S=(\Delta_m)_{m\in S}$, and for replacement values $b_S$ define $F_{S,x,y}(b_S):=\psi_e(f_{\theta_0}^{\doop(A_S:=b_S)};x,y)$. Every site in $S$ is clamped and all other computations are recomputed, so $F_{S,x,y}(a_S^0)=\psi_e(f_0;x,y)$ and $F_{S,x,y}(a_S^e)=\psi_e(P_S;x,y)$.

\begin{proposition}[Linearized recovery under joint patching]\label{prop:linear}
Suppose $F_{S,x,y}$ is twice continuously differentiable on a neighborhood of the segment $\{a_S^0+t\Delta_S:0\le t\le1\}$. Let $g_{m,S}(x,y):=\nabla_{b_m}F_{S,x,y}(a_S^0(x))$ and $H_S(t;x,y):=\nabla^2_{b_S}F_{S,x,y}(a_S^0(x)+t\Delta_S(x))$. Then
\begin{align*}
\psi_e(P_S;x,y)-\psi_e(f_0;x,y)&=\sum_{m\in S}\langle g_{m,S}(x,y),\Delta_m(x)\rangle+R_S(x,y),\\
|R_S(x,y)|&\le\tfrac12\sup_{t\in[0,1]}\|H_S(t;x,y)\|_{\mathrm{op}}\|\Delta_S(x)\|_2^2.
\end{align*}
When the summands and the remainder bound are integrable, $\mathrm{Rec}_e(S)=\sum_{m\in S}\alpha_{e,m;S}+\E[R_S(X,Y)]$, where $\alpha_{e,m;S}:=\E\langle g_{m,S}(X,Y),\Delta_m(X)\rangle$.
\end{proposition}

The first-order terms show what a single-component score measures. For $S=\{m\}$ the derivative is the usual attribution-patching derivative, which defines the singleton score $\alpha_{e,m}:=\alpha_{e,m;\{m\}}$. The derivative for a larger set can differ, because a patch on a downstream site overwrites what earlier sites send through it. If no selected site is an ancestor of another, nothing is overwritten and the joint derivatives at the base agree with the singleton derivatives. Across layers they can differ. Consider a chain $u\to v\to z$ with score $\psi=z$, in which the patch moves $u$ and $v$ by the same amount. Clamping $v$ cuts the only path from $u$ to $z$, so the joint gradient is $(0,1)$ and the set recovers the change once. Measured alone, each site has derivative $1$, so the singleton sum counts the change twice. We therefore use singleton scores to rank candidates and measure the recovery of the resulting sets directly (Appendix~\ref{app:remainder}).

\subsection{From mediator patching to localized fine-tuning}

A patch inserts specialist activations at run time, while a localized adapter updates only parameters near $S$ and has to produce the same computation through those parameters. A set that patches well therefore leaves a second question open, whether some local update can imitate the patched model. The next result separates the two requirements. Let $\calN(S)$ denote the parameter neighborhood adjacent to components in $S$ (for example the incoming/outgoing weights of selected heads or the LoRA modules attached to selected blocks), and let $\Theta(S)\subseteq\{\theta_0 + \Delta\theta:\ \mathrm{supp}(\Delta\theta)\subseteq \calN(S)\}$ be the actual admissible adaptation class. Equality holds for unrestricted local updates, while a LoRA configuration also imposes its module choices and ranks.

\begin{theorem}[Patching-to-adaptation transfer]\label{thm:transfer}
Assume the task loss is $L_\ell$-Lipschitz in the model outputs. Suppose there exists a localized adaptation $\tilde\theta\in\Theta(S)$ such that on $\calD_e$, $\E[\|f_{\tilde\theta}(X) - P_S(X)\|] \le \varepsilon_{\mathrm{approx}}$. Then
\[
\Risk_e(\tilde\theta)
\le
\Risk_e(\theta_e)
+
L_\ell
\Big(
\delta_{\mathrm{patch}}(S)
+
\varepsilon_{\mathrm{approx}}
\Big).
\]
In particular, if $S$ patches almost all of the dense specialist's behavior and the local parameter class can realize that patch, then a localized adaptation supported near $S$ achieves risk close to the dense specialist.
\end{theorem}

The proof applies the triangle inequality through the patched model, and the Lipschitz constant converts output distance into risk. The patching gap $\delta_{\mathrm{patch}}(S)$ depends on the choice of $S$, and the approximation gap $\varepsilon_{\mathrm{approx}}$ also depends on the trainable parameter class.

\begin{remark}[Scope of the transfer bound]\label{rem:scope}
The theorem is conditional on two properties, that $\Theta(S)$ can approximate the patched model (small $\varepsilon_{\mathrm{approx}}$) and that patching recovers the specialist in output space (small $\delta_{\mathrm{patch}}(S)$). It is stated in output space (logits, for cross-entropy), and we distinguish three quantities. \emph{Score recovery} $\mathrm{Rec}_e(S)$ is measured in E2. \emph{Output-space matching}, $\delta_{\mathrm{patch}}(S)$ and $\varepsilon_{\mathrm{approx}}$, consists of the theorem's hypotheses, which we do not measure here. The \emph{task performance of a trained adapter} is measured in E3 and also reflects estimation and optimization error.
\end{remark}

An empirical analogue (Remark~\ref{rem:fourterm} in Appendix~\ref{app:proofs}) replaces $\tilde\theta$ by a trained adapter $\hat\theta_S$ under uniform-deviation and empirical-suboptimality conditions, adding an estimation and an optimization term. In Section~\ref{sec:e3} we report the normalized excess task NLL of a trained adapter as an \emph{empirical adapter residual}. It is an empirical counterpart of the left-hand side of this four-term bound rather than an estimate of $\varepsilon_{\mathrm{approx}}$. Proposition~\ref{prop:restricted} in Appendix~\ref{app:restricted} adds a complementary check in parameter space, bounding the risk of keeping only the part of the dense update near $S$.

\section{Experiments}
\label{sec:experiments}

The experiments evaluate three claims. E1 tests whether activation patching recovers planted mediators under confounded activation magnitudes. E2 tests whether the specialist-vs-base gap in language models concentrates in small patchable component sets, at two model scales. E3 compares localized LoRA supports at exactly matched parameter counts. Unless stated otherwise, numbers are mean $\pm$ standard error across training seeds (five in E1, three in E2 and E3). The Pythia-1.4B check is a single seed, for which we report item-level uncertainty.

\subsection{E1: Synthetic mediator recovery}
\label{sec:e1}

\paragraph{Setup.}
E1 uses small models whose true mediators are known by construction. A 4-layer, 4-head causal transformer reads a sequence of 32 tokens and predicts one of 4 classes, and the base learns to read the class from label tokens at three fixed positions. The dense specialist is a copy of the base in which only 3 planted heads $T_e$ are fine-tuned, on data whose label tokens have moved to three other positions. Head patches act before each head's output projection. The same design is repeated in a 20-node linear SCM, in which a classifier reads $|T_e|{=}5$ planted nodes (Appendix~\ref{app:setup}). \emph{Regime A} is the planted construction, \emph{Regime B} adds parameter noise to non-planted heads, and \emph{Regime C} copies planted heads into $3$ distractor heads with output projections scaled by $0.05$, so their activations change strongly while their output effect is attenuated.

\paragraph{Results.}
We compare greedy, exhaustive, single-patch, singleton gradient-weighted ($\alpha_{e,m}$), magnitude ($\|\Delta_m\|$), and random selection (Appendix~\ref{app:e1-extras}). Table~\ref{tab:e1-jaccard} in Appendix~\ref{app:e1-extras} reports Jaccard similarity at the oracle budget $k=|T_e|$, and Figure~\ref{fig:e1-recovery} shows recovery curves. The Regime A linear SCM satisfies the hypotheses of Proposition~\ref{prop:exact}, and patching its planted set recovers the full lift ($\rho^{\mathrm{lift}}_e=1.00$). In the transformer, the fine-tuned output projections of planted heads sit downstream of the patch ($\rho^{\mathrm{lift}}_e=0.71\pm0.02$ for exhaustive search at $k=3$), so its results test support recovery under this intervention. Magnitude identifies the planted heads exactly in Regimes A and B. The discriminating case is Regime C, whose distractors change their activations strongly while their outputs are attenuated. There magnitude selection loses support recovery (Jaccard $0.48 \pm 0.15$) while greedy patching remains substantially closer to the planted mediator set ($0.80 \pm 0.12$). Recovery fractions in Regime C are $0.69 \pm 0.04$ for patching vs.\ $0.52 \pm 0.07$ for magnitude. The gap is larger in the SCM, where greedy patching achieves Jaccard $1.00$ in every regime while magnitude falls to $0.39$ under confounding.

\subsection{E2: Localization on language models}
\label{sec:e2}

\paragraph{Setup.}
E2 fine-tunes a general-purpose language model to answer questions in one subject area and asks which components carry the change. Pythia \citep{biderman2023pythia} is a family of open GPT-style models trained on a broad corpus of English text to predict the next token, with no task or subject built in. Given a prompt, it continues the text one token at a time. We use Pythia-410M, with about 410 million parameters, 24 layers, and 16 attention heads per layer, as the base generalist in E2 and E3. The three domains are question-answering benchmarks. GSM8K \citep{cobbe2021gsm8k} contains grade-school math word problems, and the model should write a worked solution ending in the numerical answer. MMLU \citep{hendrycks2021mmlu} contains multiple-choice questions in 57 subjects, from mathematics to law, and the model should give the correct option letter. PubMedQA \citep{jin2019pubmedqa} contains biomedical research questions with passages from the article's abstract as context, and the model should answer yes, no, or maybe. For each domain and seed we fine-tune all parameters on 1{,}024 training examples for three epochs to obtain a dense specialist (Appendix~\ref{app:setup}). PubMedQA training and calibration examples come from its automatically labeled PQA-A subset and test examples from its expert-labeled PQA-L subset, whose label distributions differ, with about 93\% of the PQA-A training examples labeled yes against 53--56\% of the PQA-L test items. Each domain uses 500 calibration examples (patch scores and selection only), 500 validation examples, and 500 test examples (all reported numbers), drawn separately for each seed.

\paragraph{Scored target.}
The score $\psi_e$ is the negative NLL of the target sequence, averaged over its tokens. It is teacher-forced, meaning that the model sees the prompt and the preceding reference tokens and is scored on the probability it gives the next one. The target is the reference answer followed by the EOS token, which gives 98 tokens on average for GSM8K and two tokens for MMLU and PubMedQA. The EOS token is scored like any other target token. The lift $G_e$ thus measures a change in answer-sequence likelihood, which includes answer formatting and stopping behavior.

\paragraph{Specialist quality.}
At 410 million parameters and 1{,}024 training examples, neither the base nor the specialist answers these questions reliably, so accuracy separates them only weakly. Fine-tuning does change the probability the model gives the reference answer, including its format and its stopping point, and NLL measures that change. We therefore localize on NLL. Under option-likelihood scoring (the highest-probability allowed label at the answer position), MMLU accuracy is $0.227 \pm 0.007$ for the base and $0.239 \pm 0.028$ for the specialist, against a chance level of $0.25$. PubMedQA accuracy rises from $0.404 \pm 0.009$ to $0.541 \pm 0.012$, matching the majority-label baseline ($0.540$), with the specialist predicting yes on 99.7\% of test items. On GSM8K, greedy decoding with up to 256 new tokens gives $0.020 \pm 0.003$ for the base and $0.035 \pm 0.008$ for the specialist. The NLL lift is $1.21$, $3.82$, and $8.38$ nats on GSM8K, MMLU, and PubMedQA, of which the EOS position accounts for 6\%, 89\%, and 60--63\% (Appendix~\ref{app:eos}).

\paragraph{Localization methods and metric.}
Mediator candidates consist of the $n_{\mathrm{layers}}\times n_{\mathrm{heads}}=384$ attention heads and the $24$ MLP-block outputs, for a total $|\calM|=408$. Heads are patched additively at the per-head output projection, and MLP blocks are patched by full output replacement. The final LM grid compares single-component patching (single-patch), activation magnitude $\|\Delta_m\|$, and uniformly random selection over all 408 components. Single-patch scores each component by the gain in $\psi_e$ when its specialist activation is patched into the base, averaged over the calibration set. Rankings are computed in batched groups with the other group members held at their base activations, so each MLP score is conditional on the other MLP blocks. Exact single-component scores give closely related rankings (Spearman $0.81$--$0.95$) and comparable top-$k$ recovery (Appendix~\ref{app:exact}). Performance is summarized by the area under the recovery curve (AURC) over $k\in[0,\,0.3\,|\calM|]$, which captures how efficiently each method localizes specialty within a small budget.

\paragraph{The patching ceiling.}
Embeddings and the unembedding stay at base (Figure~\ref{fig:F2_chimera}), so patching all of $\calM$ need not reproduce the specialist exactly. The ceiling is close to one in every domain ($1.004$, $0.997$, and $1.013$ on GSM8K, MMLU, and PubMedQA), and we plot $\rho^{\mathrm{patch}}_e$ for LM recovery curves.

\begin{table}[t]
\centering
\caption{\textbf{Area under the recovery curve (AURC) on three Pythia-410M language-model domains.} Mean $\pm$ standard error over 3 seeds, with higher better. AURC is computed over $k\in[0,\,0.3|\calM|]$ with the budget axis normalized by $|\calM|$, so a method that immediately reaches the patching ceiling has AURC $\approx 0.3$. Negative AURC for magnitude on GSM8K means the patch worsens the in-domain score relative to the base model before enough components accumulate.}
\label{tab:e2-aurc}
\small
\begin{tabular}{lccc}
\toprule
Method & GSM8K & MMLU & PubMedQA \\
\midrule
Single-patch (ours)   & $\mathbf{+0.250 \pm 0.003}$ & $\mathbf{+0.293 \pm 0.002}$ & $\mathbf{+0.299 \pm 0.004}$ \\
Magnitude            & $-0.118 \pm 0.012$ & $+0.225 \pm 0.018$ & $+0.287 \pm 0.002$ \\
Random               & $+0.131 \pm 0.006$ & $+0.193 \pm 0.007$ & $+0.218 \pm 0.034$ \\
\bottomrule
\end{tabular}
\end{table}

\begin{figure}[t]
\centering
\includegraphics[width=\linewidth]{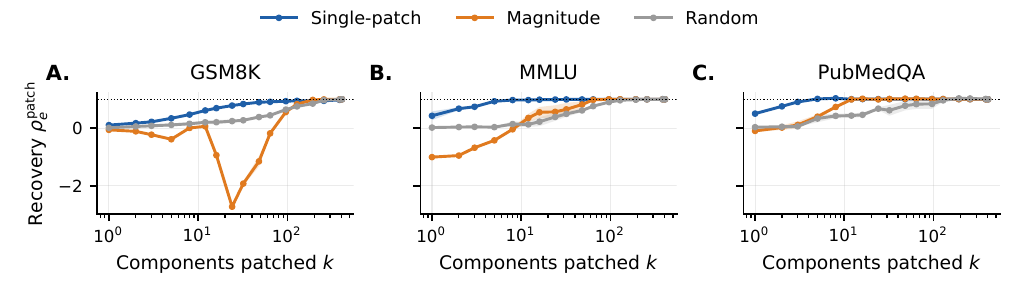}
\caption{\textbf{Recovery vs.\ number of patched components on three LM domains.} Curves show patchable-gap recovery $\rho^{\mathrm{patch}}_e$, so all methods reach $1$ at $k=|\calM|$ (log scale on the horizontal axis). Single-patch reaches half of the patchable gap at $k\approx 8$ ($2\%$ of $\calM$) on GSM8K and with one or two components on MMLU and PubMedQA, and exceeds $0.9$ at $k=5$ on both. Bands are $\pm 1$ s.e.\ over 3 seeds.}
\label{fig:e2-recovery}
\end{figure}

\paragraph{Results.}
Table~\ref{tab:e2-aurc} summarizes AURC across the three domains and three seeds, and single-patch ranks first on every (domain, seed) pair. Interference and component-type breakdowns are in Appendix~\ref{app:e2-extras}. Attention-only patching reaches $0.735$ of the GSM8K lift, against $c_e\approx 1.0$ with MLPs. Table~\ref{tab:abs-recovery} gives the curves of Figure~\ref{fig:e2-recovery} in absolute terms, and GSM8K needs sixteen components for $0.84$ of $1.21$ nats. Under exact single-component scores, MLP blocks fill three to five of the top five places in every domain and seed (Appendix~\ref{app:exact} and Figure~\ref{fig:e2-heatmap}).

\begin{table}[t]
\centering
\caption{\textbf{Absolute recovery of the dense NLL lift by the top-$k$ single-patch components.} Pythia-410M, mean $\pm$ s.e.\ over 3 seeds. Each entry is the NLL recovered by patching the top-$k$ components (nats) and, in parentheses, its fraction of the dense lift $G_e$. Values above 100\% occur because the chimera can score slightly better than the specialist (Section~\ref{sec:problem-setup}). Patching all 408 components recovers 100\%, 100\%, and 101\% of the lift.}
\label{tab:abs-recovery}
\small
\setlength{\tabcolsep}{4pt}
\begin{tabular}{@{}lcccc@{}}
\toprule
Domain & Dense lift $G_e$ & $k=3$ & $k=5$ & $k=16$ \\
\midrule
GSM8K & $1.207 \pm 0.004$ & $0.263 \pm 0.020$ (22\%) & $0.411 \pm 0.036$ (34\%) & $0.838 \pm 0.007$ (69\%) \\
MMLU & $3.820 \pm 0.023$ & $2.821 \pm 0.121$ (74\%) & $3.545 \pm 0.121$ (93\%) & $3.756 \pm 0.036$ (98\%) \\
PubMedQA & $8.379 \pm 0.157$ & $7.700 \pm 0.139$ (92\%) & $8.600 \pm 0.078$ (103\%) & $8.554 \pm 0.083$ (102\%) \\
\bottomrule
\end{tabular}
\end{table}

\paragraph{Ranking stability.}
With the specialist fixed, rankings correlate at $0.978$ across disjoint calibration halves (1.4B) and at $0.71$--$0.99$ across prompt templates. Accuracy-based rankings are moderately related to NLL-based ones on PubMedQA and unrelated on MMLU, where accuracy is at chance (Appendix~\ref{app:stability}).

\subsection{Scale check: Pythia-1.4B}
\label{sec:scale}

To test whether the concentration survives a larger backbone while holding the model family and specialist construction fixed, we repeated the PubMedQA protocol with Pythia-1.4B, the member of the same family with about 1.4 billion parameters (24 layers, 16 heads, so again $|\calM|=408$), using dense fine-tuning on the same 1{,}024 training examples for three epochs, 256 calibration examples, and the same 500 test examples (seed 0). On the test set, NLL falls from $6.767$ to $1.458$, a paired lift of $5.309$ nats (item-level s.e.\ $0.065$, 95\% paired bootstrap interval $[5.201, 5.436]$). Table~\ref{tab:scale} shows that single-patch recovers 76\% of the lift with three components and 94\% with five, while magnitude selection moves the model away from the specialist and random selection recovers almost nothing. The five top-ranked components are all MLP blocks. With exact single-component scores the top five are again all MLP blocks and recover 88\% (Appendix~\ref{app:exact}). As at 410M, the EOS position carries 55\% of the lift (Appendix~\ref{app:eos}).

\begin{table}[t]
\centering
\caption{\textbf{Pythia-1.4B PubMedQA, one seed.} NLL recovered (nats) and fraction of the $5.309$-nat dense lift for the top-$k$ components of each ranking, on 500 test items. Patching all 408 components recovers $5.322$ nats (100.2\%).}
\label{tab:scale}
\small
\begin{tabular}{@{}lccc@{}}
\toprule
Ranking & $k=3$ & $k=5$ & $k=16$ \\
\midrule
Single-patch & $4.030$ (75.9\%) & $4.975$ (93.7\%) & $5.700$ (107.4\%) \\
Magnitude & $-4.226$ ($-79.6$\%) & $-3.774$ ($-71.1$\%) & $-0.872$ ($-16.4$\%) \\
Random & $0.188$ (3.5\%) & $0.209$ (3.9\%) & $2.343$ (44.1\%) \\
\bottomrule
\end{tabular}
\end{table}

\subsection{E3: Localized adaptation at matched parameter budgets}
\label{sec:e3}

E3 moves from patching to training. Instead of fine-tuning the whole model, we train a small adapter on a few blocks of Pythia-410M with the same domain data, and ask whether blocks chosen from the patching ranking work better than other blocks at the same parameter count.

\paragraph{Setup.}
LoRA \citep{hu2021lora} freezes the model's weights and trains a small low-rank update on selected modules. An adapter is set by its support, the $k$ of $N{=}24$ transformer blocks it may update, and by the rank $r$ of the update. Every adapter updates the attention and MLP projections of its blocks, and the five methods differ only in the support. \emph{Patching-localized} takes the $k$ blocks that contain the highest-ranked heads in the attention-head entries of the E2 single-patch ranking. \emph{Random-localized} takes $k$ blocks at random. \emph{Dense-update-norm} takes the $k$ blocks whose weights moved most under dense fine-tuning ($\|\theta_e-\theta_0\|$). \emph{Domain-probe} is a training-free selector that ranks blocks by activation separation between in-domain and off-domain batches \citep[inspired by][]{yin2024lofit}. \emph{Full LoRA} adapts all $N$ blocks at the scaled rank $r'$ that gives the same budget. Parameter budgets match exactly, so in each of the nine $(k/N, r)$ cells all 45 adapters (five methods, three domains, three seeds) have identical trainable-parameter counts, from $393{,}216$ at $(0.25, 4)$ to $4{,}718{,}592$ at $(0.75, 16)$. We train on 256 examples for 5 epochs over $k/N{\in}\{0.25,0.5,0.75\}$ and $r{\in}\{4,8,16\}$, and compare supports by test NLL.

\begin{table}[t]
\centering
\caption{\textbf{Localized adapters at matched parameter budgets.} \emph{Grid NLL} is in-domain test NLL averaged over the nine $(k/N, r)$ cells within each training seed (mean $\pm$ s.e.\ across 3 seeds, lowest in bold). \emph{Residual} is the empirical adapter residual at the method's lowest-NLL cell, in percent. Dense specialist NLL is $1.051$, $0.714$, and $1.350$, and base NLL is $2.173$, $4.527$, and $9.726$.}
\label{tab:e3-nll}
\small
\setlength{\tabcolsep}{2.7pt}
\begin{tabular}{@{}lcccccc@{}}
\toprule
 & \multicolumn{2}{c}{GSM8K} & \multicolumn{2}{c}{MMLU} & \multicolumn{2}{c}{PubMedQA} \\
\cmidrule(lr){2-3}\cmidrule(lr){4-5}\cmidrule(l){6-7}
Method & Grid NLL & Residual & Grid NLL & Residual & Grid NLL & Residual \\
\midrule
Patching-localized & $1.3098 \pm 0.0071$ & 12.8 & $\mathbf{0.7268 \pm 0.0079}$ & 0.06 & $\mathbf{1.178 \pm 0.146}$ & $-3.4$ \\
Full LoRA & $\mathbf{1.2272 \pm 0.0019}$ & 11.7 & $0.7271 \pm 0.0087$ & 0.02 & $1.296 \pm 0.191$ & $-1.8$ \\
Random-localized & $1.2776 \pm 0.0135$ & 12.8 & $0.7298 \pm 0.0083$ & $-0.04$ & $1.192 \pm 0.146$ & $-3.7$ \\
Dense-update-norm & $1.2777 \pm 0.0044$ & 11.8 & $0.7390 \pm 0.0080$ & 0.19 & $1.226 \pm 0.148$ & $-2.6$ \\
Domain-probe & $1.2985 \pm 0.0016$ & 13.2 & $0.7325 \pm 0.0081$ & 0.05 & $1.222 \pm 0.177$ & $-3.1$ \\
\bottomrule
\end{tabular}
\end{table}

\paragraph{Results.}
Full LoRA has the lowest grid-averaged test NLL on GSM8K (Table~\ref{tab:e3-nll}), below patching-localized in each seed ($1.227$ vs.\ $1.310$). On MMLU and PubMedQA, patching-localized has the lowest mean and is below random-localized in each seed, by margins smaller than the across-seed variation. Table~\ref{tab:e3-nll} also reports each method at its own lowest-NLL cell as an \emph{empirical adapter residual}, the fraction of the dense NLL gain that the adapter leaves unrecovered, $(\mathrm{NLL}_{\text{adapter}}-\mathrm{NLL}_{\text{dense}})/(\mathrm{NLL}_{\text{base}}-\mathrm{NLL}_{\text{dense}})$. On GSM8K every method leaves 11.7--13.2\% of the gain, and every method's best cell is the largest budget in the grid, although patching all 408 components reproduces the specialist's GSM8K score ($c_e=1.004$). On MMLU every residual is below 0.2\%, and on PubMedQA every method's best cell reaches lower test NLL than the dense specialist. The similar GSM8K residuals are compatible with a limitation shared by block-level LoRA under this training budget, and separating its sources is future work. Block selection uses head scores while MLP blocks are among the highest-ranked components in E2, and LoRA adapts whole blocks while E2 ranks individual components, which leaves room for supports designed around the full ranking. Appendix~\ref{app:e3-extras} adds block-overlap, tiny-budget, and head-level adapter comparisons.

\section{Discussion and Conclusion}
\label{sec:discussion}

Our central object is the specialist mediator, the internal components through which dense fine-tuning changes the model's in-domain score. The results separate three levels of evidence. \emph{Score recovery}, whether patching a few components recovers the specialist's task score, is what E2 measures. \emph{Output-space matching}, whether the patched model's outputs match the specialist's ($\delta_{\mathrm{patch}}$) and whether a local class can match the patched model ($\varepsilon_{\mathrm{approx}}$), is the level at which Theorem~\ref{thm:transfer} is stated. \emph{Trained-adapter performance}, whether an adapter on the localized support reaches the specialist's task risk, is what E3 measures. Successful activation localization does not by itself yield a better localized adapter.

This separation gives a scoped order of checks for anyone who wants to use localization before sparse adaptation. A low full-set reference signals a mismatch worth investigating, possibly a candidate set that omits changed components, though a smaller set can still patch well.  If a small set patches well but adapters on its support fit poorly, the open questions concern the adapter's capacity, module support, and training signal. If both succeed, the set is a candidate sparse-adaptation target.

\paragraph{Limitations and outlook.}
The language-model results cover one model family at two scales, with one seed and one domain at 1.4B. Natural next steps are other model families and PEFT-trained specialists at larger scale, answer-only scores that exclude EOS, and per-token $\mathrm{KL}(f_e\,\|\,P_S)$ and $\mathrm{KL}(P_S\,\|\,f_{\hat\theta_S})$ with adapters distilled to the patched model, which would test agreement with the patch more directly. Cheaper specialization also lowers the barrier for harmful specialists with subtler weight drift, and localization itself can serve as a detection diagnostic.

\begin{ack}
This work received no third-party funding. The author declares no competing interests.
\end{ack}

\bibliographystyle{unsrtnat}
\bibliography{references}

\clearpage
\appendix

\section{Full proofs and the four-term empirical bound}
\label{app:proofs}

This appendix proves the results of Section~\ref{sec:theory}. We first give the empirical version of the transfer bound mentioned in Section~\ref{sec:theory}, which replaces the comparator of Theorem~\ref{thm:transfer} by a trained adapter and so adds an estimation term and an optimization term.

\begin{remark}[Empirical learning decomposition]\label{rem:fourterm}
Let $\widehat\Risk_{e,n}$ be empirical task risk, let $\hat\theta_S\in\Theta(S)$ be a trained adapter, and let $\tilde\theta\in\Theta(S)$ be a comparator satisfying Theorem~\ref{thm:transfer}. Suppose, on an event of interest,
\[
\sup_{\theta\in\Theta(S)}|\Risk_e(\theta)-\widehat\Risk_{e,n}(\theta)|\le\eta_S,
\qquad
\widehat\Risk_{e,n}(\hat\theta_S)\le\inf_{\theta\in\Theta(S)}\widehat\Risk_{e,n}(\theta)+\varepsilon_{\mathrm{opt}}.
\]
On that event,
\[
\Risk_e(\hat\theta_S)-\Risk_e(\theta_e)\le L_\ell\big(\delta_{\mathrm{patch}}(S)+\varepsilon_{\mathrm{approx}}\big)+2\eta_S+\varepsilon_{\mathrm{opt}}.
\]
Indeed, $\Risk_e(\hat\theta_S)\le\widehat\Risk_{e,n}(\hat\theta_S)+\eta_S\le\widehat\Risk_{e,n}(\tilde\theta)+\varepsilon_{\mathrm{opt}}+\eta_S\le\Risk_e(\tilde\theta)+\varepsilon_{\mathrm{opt}}+2\eta_S$, and Theorem~\ref{thm:transfer} bounds the last risk. This is the four-term decomposition with estimation term $\varepsilon_{\mathrm{est}}=2\eta_S$. If the class is selected using the same data, the deviation event must hold simultaneously over the possible selected classes.
\end{remark}

\subsection*{Proof of Proposition~\ref{prop:exact} (exact mediator sufficiency)}

Let $\mathcal{V}=\{m_1,\ldots,m_M\}\supseteq\calM$ be the set of all nodes of the finite computation graph, ordered so that $m_i$ is an ancestor of $m_j$ only if $i<j$ (topological order). For any node $m_i$, write $\mathrm{pa}(m_i)$ for its parent set and let $g_i^0$ and $g_i^e$ denote the structural equations of $m_i$ in the base model and specialist respectively. By hypothesis, $g_i^0 = g_i^e$ for all $m_i \notin T_e$.

We prove by induction on $i$ that the activation of the patched model $P_{T_e}$ equals that of the specialist at every node. \emph{Base case.} Consider $m_1$. If $m_1 \in T_e$, the patching operator replaces $a_{m_1}^0(x)$ with $a_{m_1}^e(x)$, giving equality. If $m_1 \notin T_e$, then $g_1^0 = g_1^e$ and $m_1$ has no parents, so $a_{m_1}^0(x) = g_1^0(x) = g_1^e(x) = a_{m_1}^e(x)$. \emph{Inductive step.} Suppose for all $j<i$ that the patched model's activation at $m_j$ equals $a_{m_j}^e(x)$. If $m_i\in T_e$, patching forces equality. If $m_i\notin T_e$, $g_i^0=g_i^e$ and parent activations agree by induction, so $a_{m_i}^{P_{T_e}}(x)=g_i^0(a_{\mathrm{pa}(m_i)}^e(x))=g_i^e(a_{\mathrm{pa}(m_i)}^e(x))=a_{m_i}^e(x)$. The model output is itself a node of $\mathcal{V}$, so $P_{T_e}(x)=f_e(x)$ for every $x$, and the $\rho^{\mathrm{lift}}_e$ and $\mathrm{Int}_e$ claims follow by substitution.\qed

\subsection*{Proof of Proposition~\ref{prop:linear} (linearized recovery under joint patching)}

Fix $(x,y)$ and $S$, and put $h(t)=F_{S,x,y}(a_S^0+t\Delta_S)$. Clamping at the base values reproduces the base forward pass, so $h(0)=\psi_e(f_0;x,y)$, and clamping at the specialist values gives $h(1)=\psi_e(P_S;x,y)$. The chain rule gives $h'(0)=\sum_{m\in S}\langle g_{m,S},\Delta_m\rangle$ and $h''(t)=\Delta_S^\top H_S(t;x,y)\Delta_S$. Taylor's formula with integral remainder gives
\[
h(1)-h(0)=h'(0)+\int_0^1(1-t)\,\Delta_S^\top H_S(t;x,y)\Delta_S\,dt.
\]
The operator-norm inequality and $\int_0^1(1-t)\,dt=\tfrac12$ give the stated bound, and integrability permits taking expectations. For a singleton there are no other clamps. If selected sites have no ancestor relations, clamping another selected site at its base value removes no directed path from $m$ to the output and changes no base value, so the derivative at the base is the singleton derivative.\qed

\subsection*{Proof of Theorem~\ref{thm:transfer} (patching-to-adaptation transfer)}

Write $\ell_e(f(x),y)$ for the per-example loss, so $\Risk_e(\theta) = \E_{(x,y)\sim\calD_e}[\ell_e(f_\theta(x),y)]$. By $L_\ell$-Lipschitzness of $\ell_e$ in its first argument, $|\ell_e(f_{\tilde\theta}(x),y) - \ell_e(f_{\theta_e}(x),y)| \le L_\ell \|f_{\tilde\theta}(x)-f_{\theta_e}(x)\|$. By the triangle inequality, $\|f_{\tilde\theta}(x)-f_{\theta_e}(x)\|\le \|f_{\tilde\theta}(x)-P_S(x)\| + \|P_S(x)-f_{\theta_e}(x)\|$. Taking expectations,
\begin{align*}
\Risk_e(\tilde\theta) - \Risk_e(\theta_e)
&\le L_\ell\,\E\big[\|f_{\tilde\theta}(X)-f_{\theta_e}(X)\|\big] \\
&\le L_\ell\Big(\E\big[\|f_{\tilde\theta}(X)-P_S(X)\|\big] + \E\big[\|P_S(X)-f_{\theta_e}(X)\|\big]\Big) \\
&\le L_\ell\big(\varepsilon_{\mathrm{approx}} + \delta_{\mathrm{patch}}(S)\big),
\end{align*}
where the last step uses the hypothesis on $\tilde\theta$. \qed

For token-level cross-entropy with model outputs taken as logits, the Lipschitz condition holds with $L_\ell=\sqrt{2}$ in the Euclidean norm, since the gradient of the loss with respect to the logits is the difference between the predicted distribution and a one-hot vector.

\section{A parameter-space bound}
\label{app:restricted}

Theorem~\ref{thm:transfer} reasons through the patched model. A simpler question asks what happens if we keep the part of the dense update near $S$ and discard the rest. When the risk is smooth around the dense specialist, the cost depends on the size of the discarded part, so it is small when dense fine-tuning moved little outside $\calN(S)$.

\begin{proposition}[Restricted-update bound]\label{prop:restricted}
Assume the population risk $\Risk_e$ is $\beta_e$-smooth in parameters. Let $\Delta\theta^{\mathrm{full}}:=\theta_e-\theta_0$ be the dense specialist update, decomposed as $\Delta\theta^{\mathrm{full}} = \Delta\theta_S + \Delta\theta_{\bar S}$, where $\Delta\theta_S$ is supported on $\calN(S)$ and $\Delta\theta_{\bar S}$ on its complement. Then
\[
\Risk_e(\theta_0+\Delta\theta_S)
-
\Risk_e(\theta_e)
\le
\left\langle
\nabla \Risk_e(\theta_e),\, -\Delta\theta_{\bar S}
\right\rangle
+
\frac{\beta_e}{2}\|\Delta\theta_{\bar S}\|_2^2.
\]
If $\|\nabla\Risk_e(\theta_e)\|_2\le\gamma_e$, the right-hand side is at most $\gamma_e\|\Delta\theta_{\bar S}\|_2+\tfrac{\beta_e}{2}\|\Delta\theta_{\bar S}\|_2^2$.
\end{proposition}

The truncated update keeps the dense update's values, so it need not belong to a low-rank adapter class.

\subsection*{Proof of Proposition~\ref{prop:restricted} (restricted-update bound)}

Since $\theta_0+\Delta\theta_S = \theta_e - \Delta\theta_{\bar S}$, $\beta_e$-smoothness of $\Risk_e$ gives the quadratic upper bound $\Risk_e(\theta_e-\Delta\theta_{\bar S})\le \Risk_e(\theta_e)+\langle\nabla\Risk_e(\theta_e),-\Delta\theta_{\bar S}\rangle+\tfrac{\beta_e}{2}\|\Delta\theta_{\bar S}\|_2^2$. Cauchy--Schwarz gives the quantified form.\qed

\section{Remainder bound for the linearized decomposition}
\label{app:remainder}

The remainder in Proposition~\ref{prop:linear} measures how far the first-order sum is from the recovery of the set, and blocks of the Hessian show where that error comes from. With block norms $h_{mn}:=\sup_{t\in[0,1]}\|[H_S(t)]_{mn}\|_{\mathrm{op}}$,
\[
|R_S|\le\frac12\sum_{m,n\in S}h_{mn}\|\Delta_m\|_2\|\Delta_n\|_2.
\]
The off-diagonal blocks measure how much two patched components interact, and the diagonal blocks measure the curvature of each component's own effect. The linear approximation is useful when both are small at the size of the activation change.

\begin{remark}[Selection heuristics and attribution patching]\label{rem:heuristics}
The singleton coefficients $\alpha_{e,m}$ are the familiar attribution-patching scores \citep{nanda2023attribution, syed2024attribution, kramar2024atp}. A top-$k$ guarantee for recovery would require uniform control of the departure of $\mathrm{Rec}_e$ from a set-independent additive score. The coefficients are structurally analogous to curvature-aware importance scores for structured pruning, which expand a replacement risk around a hard intervention (replacing a unit by a constant) rather than around an activation swap \citep{asiaee2026efficient}. We apply this standard expansion to the specialist-vs-generalist recovery problem and connect it to localized adaptation via Theorem~\ref{thm:transfer}.
\end{remark}

\section{Detailed experimental setup}
\label{app:setup}

This appendix gives the details needed to reproduce the experiments of Section~\ref{sec:experiments}.

\paragraph{Synthetic transformer (E1).}
The architecture is a 4-layer causal transformer with 4 heads per layer, $d_{\mathrm{model}}=128$, $d_{\mathrm{head}}=32$, $d_{\mathrm{ff}}=512$, vocabulary size 64, sequence length 32, 4 classes. Each sequence starts with a classification token, and its other tokens are random except at six signal positions. In the base data, positions 1--3 hold the label's token with probability $0.8$ (otherwise a wrong class token) and positions 4--6 are random. In the specialist data, positions 1--3 hold the label's token with probability $0.2$ and positions 4--6 with probability $0.9$. The base trains on all parameters for 250 steps, and the specialist is a copy in which the query, key, value, and output projections of 3 planted heads train for 220 steps. Each head's cached and patched activation is taken before its output projection. In Regime C, copying and attenuating happen after specialist training, and the distractors keep a small effect on the output.

\paragraph{Linear SCM (E1).}
A deterministic lower-triangular linear model over $d=20$ nodes reads six Gaussian input features, two of which carry the label signal. Base input weights are drawn from $\mathcal{N}(0,0.08^2)$ and node-to-node weights from $\mathcal{N}(0,0.03^2)$, and the classifier reads 5 planted mediator nodes. The specialist adds to the planted nodes' input weights and to their weight on the preceding node. Regime B adds noise to the structural equations of non-planted nodes. Regime C makes 3 distractor nodes depend strongly on planted nodes, with zero classifier weight.

\paragraph{Output representation for the transfer bound.}
For token-level cross-entropy, Theorem~\ref{thm:transfer} takes outputs to be vocabulary logits with the Euclidean norm. In the teacher-forced application, draw an example and then a scored target position uniformly within it. Here $X$ is the prompt and reference prefix and $Y$ the next target token. The expected token loss then equals the expected per-example mean target-token NLL, and all three models are evaluated on the same context. We use logits because probability-space NLL has no global Lipschitz constant. The reported GSM8K values follow slightly different token weighting, per batch in E2 and over all test tokens in E3. The two-token targets of MMLU and PubMedQA are unaffected.

\paragraph{Language model fine-tuning (E2--E3).}
The base model is Pythia-410M \citep{biderman2023pythia} (24 layers, 16 attention heads, $d_{\mathrm{model}}{=}1024$). Dense specialists are obtained by full fine-tuning for 3 epochs on 1{,}024 training examples with AdamW (learning rate $2\times 10^{-5}$, weight decay $0.01$), a cosine schedule with linear warm-up over the first 10\% of the 384 updates, per-device batch size 2, and 4 gradient-accumulation steps (effective batch size 8). Training examples are the first 1{,}024 of a seed-specific shuffle of the GSM8K training split (after its calibration and validation examples), the MMLU auxiliary training split, and the PubMedQA PQA-A (artificial) subset (after its calibration examples). Calibration examples (500) come from the GSM8K training split, the MMLU validation split, and PQA-A. Validation examples (500) come from the GSM8K training split, the MMLU validation split, and PQA-L. Test examples (500) come from the GSM8K test split, the MMLU test split, and PQA-L. Inputs are truncated from the left to 256 tokens so that the answer cue is kept. Patching, recovery curves, and E3 evaluation run in 16-bit floating point (float16).

\paragraph{Prompt templates.}
GSM8K: ``Solve the grade-school math problem.\textbackslash nQuestion: \{question\}\textbackslash nAnswer:'', target the reference solution, which ends in ``\#\#\#\# \{answer\}''. PubMedQA: ``Read the biomedical abstract and answer yes, no, or maybe.\textbackslash nQuestion: \{question\}\textbackslash nContext: \{first two context passages\}\textbackslash nAnswer:'', target the label. MMLU: ``Answer the multiple-choice question by giving the correct option letter.\textbackslash nSubject: \{subject\}\textbackslash nQuestion: \{question\}\textbackslash nChoices:\textbackslash nA. \ldots\textbackslash nAnswer:'', target the option letter. Each target is preceded by a space and followed by EOS.

\paragraph{Accuracy evaluation.}
MMLU and PubMedQA accuracy uses option-likelihood scoring, in which the prediction is the allowed label (A--D, or yes/no/maybe) with the highest next-token probability after the answer cue, with prompts left-truncated to their final 256 tokens. GSM8K accuracy uses greedy decoding for up to 256 new tokens, and the prediction is the number after ``\#\#\#\#'' or, if absent, the last number in the completion. The specialist ends 97\% of its completions with ``\#\#\#\#'', the base none.

\paragraph{Localized LoRA (E3).}
For a set of selected blocks, LoRA modules are attached to the fused query-key-value projection, the attention output projection, and both MLP projections of each block, so each adapted block contributes $16{,}384\,r$ trainable parameters. Training uses AdamW with learning rate $10^{-4}$ and weight decay $0.01$, a cosine schedule with 10\% warm-up, 5 epochs, batch size 8, and 4 gradient-accumulation steps on the first 256 of the specialist's 1{,}024 training examples. The tiny-budget runs in Appendix~\ref{app:e3-extras} use all 1{,}024. Full LoRA adapts all 24 blocks at the rank $r'=r\,k/N$ that matches the localized budget exactly. The domain-probe selector uses the mean activation separation between in-domain and off-domain calibration batches, a coarser, training-free variant inspired by LoFiT \citep{yin2024lofit}.

\paragraph{Pythia-1.4B scale check.}
Pythia-1.4B ($d_{\mathrm{model}}{=}2048$, 24 layers, 16 heads) is fine-tuned densely in bfloat16 with gradient checkpointing on the same 1{,}024 PubMedQA examples for three epochs (learning rate $2\times 10^{-5}$), with seed-0 splits. Single-patch scores use 256 calibration examples, and recovery uses the 500 test examples. Training took 5 minutes and single-patch scoring 15 minutes on one 24 GB GPU.

\section{Extended related work: causal abstraction, pruning, and modularity}
\label{app:related-broader}

\input{related_work_appendix.tex}

\section{E1 methods, support recovery, and recovery curves}
\label{app:e1-extras}

\paragraph{Localization methods.}
This appendix gives the methods and full results behind Section~\ref{sec:e1}. We compare (a) greedy forward selection (add the component with the largest marginal recovery gain); (b) exhaustive patching search over all size-$k$ subsets, reported up to $k=|T_e|$ as an oracle baseline; (c) top-$k$ by individual single-patch recovery; (d) gradient-weighted ranking via the singleton score $\alpha_{e,m}$ (transformer only); (e) magnitude ranking by $\|\Delta_m\|$; and (f) uniform random selection.

\begin{table}[h]
\centering
\caption{\textbf{Support recovery (Jaccard) at $k=|T_e|$.} Jaccard similarity between the selected set and the planted set $T_e$. Transformer model, mean $\pm$ s.e.\ over 5 seeds. Bold marks the best per regime. In Regimes A--B, magnitude perfectly identifies the planted heads. In Regime C, magnitude drops to $0.48$ while greedy patching maintains $0.80$.}
\label{tab:e1-jaccard}
\small
\begin{tabular}{lccc}
\toprule
Method & Regime A & Regime B & Regime C \\
\midrule
Greedy patching & $0.90 \pm 0.10$ & $0.90 \pm 0.10$ & $\mathbf{0.80 \pm 0.12}$ \\
Exhaustive patching & $0.90 \pm 0.10$ & $0.90 \pm 0.10$ & $\mathbf{0.80 \pm 0.12}$ \\
Gradient-weighted & $0.70 \pm 0.12$ & $0.80 \pm 0.12$ & $0.74 \pm 0.17$ \\
Single-patch & $0.60 \pm 0.10$ & $0.80 \pm 0.12$ & $0.64 \pm 0.16$ \\
Magnitude & $\mathbf{1.00 \pm 0.00}$ & $\mathbf{1.00 \pm 0.00}$ & $0.48 \pm 0.15$ \\
Random & $0.08 \pm 0.05$ & $0.08 \pm 0.05$ & $0.08 \pm 0.05$ \\
\bottomrule
\end{tabular}
\end{table}

\begin{figure}[h]
\centering
\includegraphics[width=0.82\linewidth]{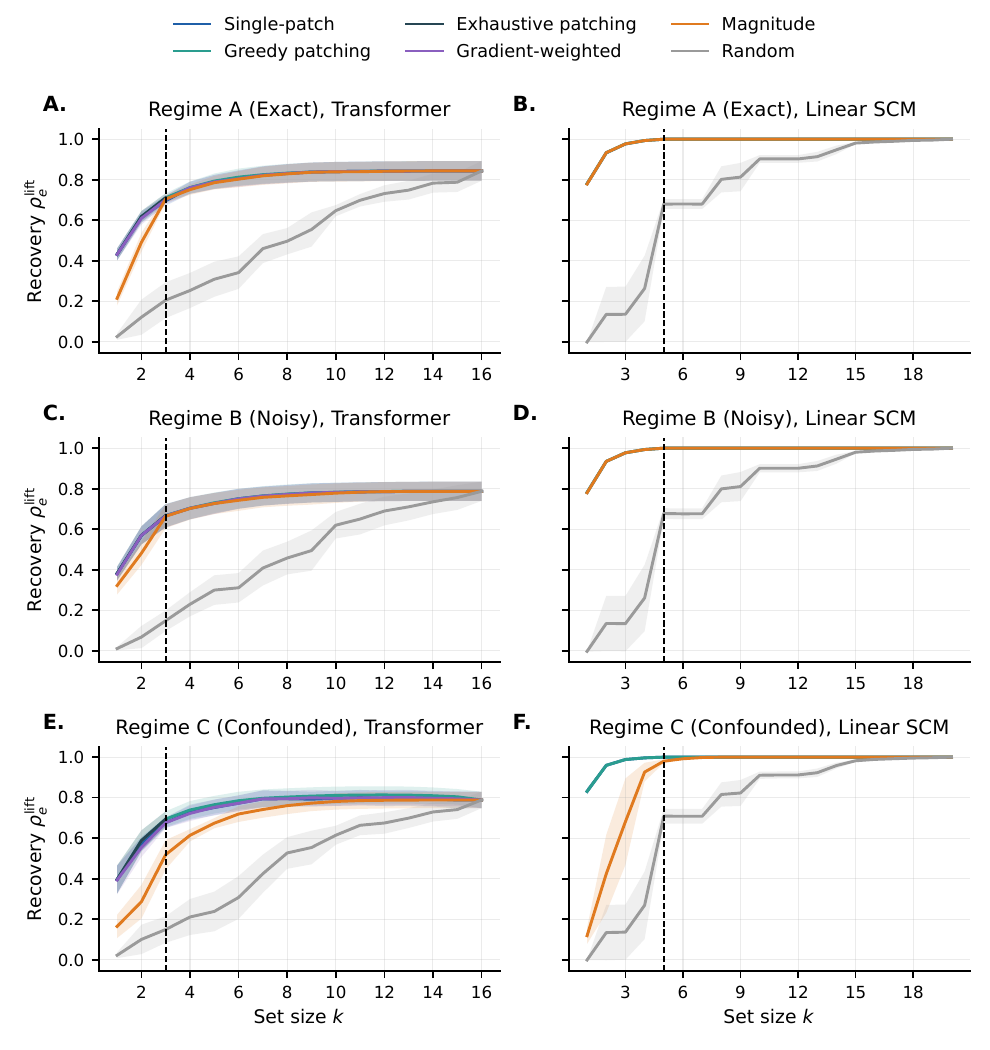}
\caption{\textbf{Synthetic mediator recovery across three regimes.} Dense-lift recovery $\rho^{\mathrm{lift}}_e(S)$ vs.\ mediator set size $k$. The left column is the transformer and the right column the linear SCM. Rows show Regime A (exact), B (noisy), and C (confounded). Gradient-weighted curves are shown for the transformer only. Bands are $\pm 1$ s.e.\ over 5 seeds. The dashed line marks $k=|T_e|$.}
\label{fig:e1-recovery}
\end{figure}

\section{E2 extras: ceiling, interference frontier, selection heatmap}
\label{app:e2-extras}

\begin{figure}[h]
  \centering
  \begin{subfigure}[t]{0.42\linewidth}
    \centering
    \includegraphics[width=\linewidth]{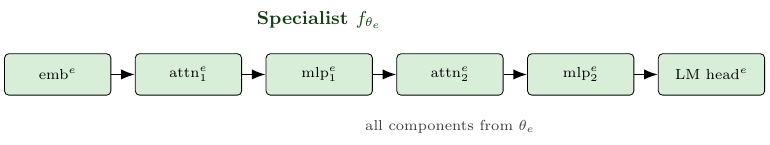}
    \caption{Specialist forward pass.}
    \label{fig:F2a_specialist}
  \end{subfigure}\hfill
  \begin{subfigure}[t]{0.42\linewidth}
    \centering
    \includegraphics[width=\linewidth]{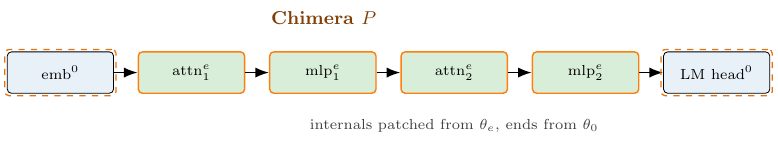}
    \caption{All internal components patched.}
    \label{fig:F2b_chimera}
  \end{subfigure}
  \caption{\textbf{Why full internal patching can still be a chimera.} If $\calM$ omits embeddings, final layer norm, or the unembedding, then patching every component in $\calM$ puts specialist internals between base endpoints. This motivates the patching ceiling $c_e$ and the patchable-gap normalization $\rho^{\mathrm{patch}}_e$.}
  \label{fig:F2_chimera}
\end{figure}

\begin{figure}[h]
\centering
\includegraphics[width=\linewidth]{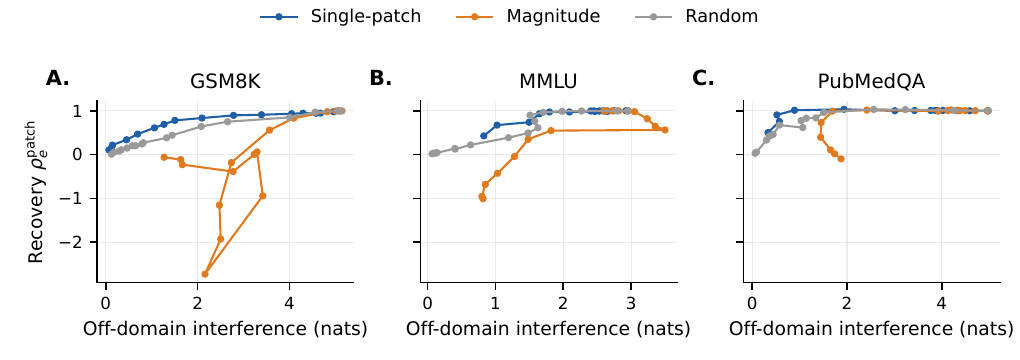}
\caption{\textbf{Recovery--interference trade-off.} Each point is one budget $k$ of one ranking, averaged over 3 seeds, with points connected in order of increasing $k$. Single-patch traces a higher-recovery, lower-interference frontier than magnitude or random. Interference is the absolute change in NLL on the two non-target domains, averaged.}
\label{fig:e2-tradeoff}
\end{figure}

\begin{figure}[h]
\centering
\includegraphics[width=\linewidth]{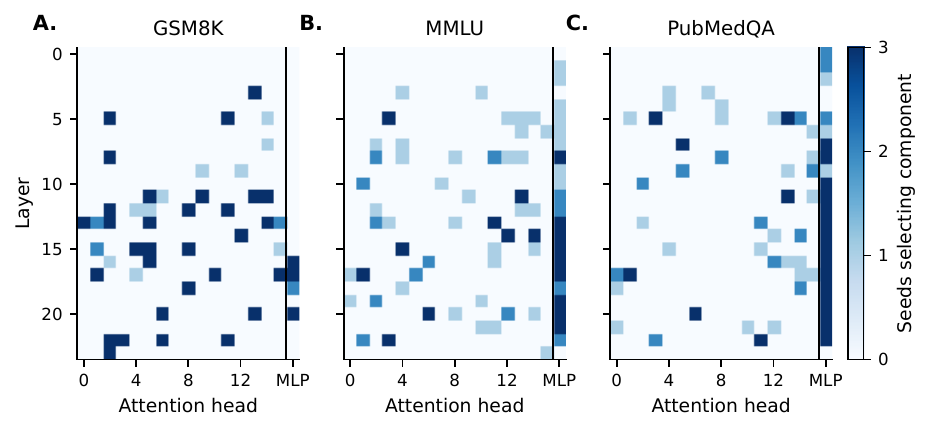}
\caption{\textbf{Where the top-ranked components sit.} For each domain, the number of seeds (out of 3) in which a component is among the 40 highest single-patch scores (10\% of $\calM$). Columns 0--15 are attention heads, and the last column is the MLP block of each layer. MLP blocks are frequent among the top components on MMLU and PubMedQA.}
\label{fig:e2-heatmap}
\end{figure}

This appendix collects results behind Section~\ref{sec:e2}. Figure~\ref{fig:F2_chimera} shows why patching every component in $\calM$ differs from running the specialist. Table~\ref{tab:e2-ceiling} breaks the patching ceiling down by component type. Figure~\ref{fig:e2-tradeoff} compares in-domain recovery with off-domain interference, and Figure~\ref{fig:e2-heatmap} shows where the highest-ranked components sit.

\begin{table}[h]
\centering
\caption{\textbf{Patching ceiling and component-type contribution.} Dense-lift recovery $\mathrm{Rec}_e(S)/G_e$ for each component scope, mean $\pm$ s.e.\ over 3 seeds. The Attention+MLP row is the ceiling $c_e$. Attention-only captures most of the dense lift on MMLU and PubMedQA but only $\approx 73\%$ on GSM8K, where MLPs contribute substantially. Values above $1$ mean the patched model scores slightly better than the specialist, and negative values mean the patch lowers the score below the base.}
\label{tab:e2-ceiling}
\small
\begin{tabular}{lccc}
\toprule
Component scope & GSM8K & MMLU & PubMedQA \\
\midrule
Attention only ($384$ heads)        & $0.735 \pm 0.006$ & $0.937 \pm 0.036$ & $1.034 \pm 0.005$ \\
MLPs only ($24$ blocks)             & $-1.607 \pm 0.110$ & $0.745 \pm 0.123$ & $1.035 \pm 0.015$ \\
Attention + MLP (full $\calM$)       & $1.004 \pm 0.001$ & $0.997 \pm 0.000$ & $1.013 \pm 0.001$ \\
\bottomrule
\end{tabular}
\end{table}

\section{The answer-sequence target and the EOS position}
\label{app:eos}

The scored target in Section~\ref{sec:e2} ends in the EOS token, so part of the lift can come from the model learning to stop after its answer. Table~\ref{tab:eos} splits the dense NLL lift into the contribution of the answer tokens and of the final EOS token. The evaluation reuses the E2 test split, precision (float16 at 410M, bfloat16 autocast at 1.4B), batching, and aggregation (per-batch token mean, averaged over batches), and reproduces the reported dense lift of every run to within $2\times10^{-4}$ nats. Because the lift is an average over target tokens, the two parts add up to the lift. The last two columns compare the specialist's NLL on the answer token alone with the entropy of the test-set label frequencies, which is the NLL of a constant predictor that outputs those frequencies.

\begin{table}[h]
\centering
\caption{\textbf{Composition of the dense NLL lift.} EOS share is the fraction of the lift at the EOS position, answer-token NLL is the specialist's NLL on the answer token(s) alone, and label entropy is the entropy of the test-set label frequencies. Pythia-410M rows are mean (range) over 3 seeds, and the Pythia-1.4B row is one seed.}
\label{tab:eos}
\small
\setlength{\tabcolsep}{4pt}
\begin{tabular}{@{}llcccc@{}}
\toprule
Model & Domain & Dense lift (nats) & EOS share & Answer-token NLL & Label entropy \\
\midrule
410M & GSM8K & $1.207$ & 6.4\% (6.4--6.5) & $1.026$ (per token) & -- \\
410M & MMLU & $3.820$ & 89.2\% (89.0--89.2) & $1.43$ (1.39--1.48) & $1.38$ \\
410M & PubMedQA & $8.379$ & 61.6\% (59.5--62.9) & $2.70$ (2.08--3.01) & $0.93$ \\
1.4B & PubMedQA & $5.309$ & 55.4\% & $2.91$ & $0.94$ \\
\bottomrule
\end{tabular}
\end{table}

On MMLU and PubMedQA the specialists assign almost all probability to EOS after the answer (specialist EOS NLL below $0.004$ nats), whereas the base model does not. This is the largest single source of the lift. On the answer token, the specialists' NLL stays at or above the label-frequency entropy, and the PubMedQA specialist predicts yes with high confidence. The localized NLL lift on these two domains therefore reflects answer formatting, stopping, and the training label prior.

\section{Exact single-component scores}
\label{app:exact}

The reported single-patch rankings (Section~\ref{sec:e2}) score components in groups. One expanded forward pass patches each component of a group in its own copy of the batch while the other group members keep their base activations. When a group spans layers, a component's score excludes effects that pass through the later members of its group. This appendix compares those rankings with exact scores. In the notation of Proposition~\ref{prop:linear}, it is a coordinate-wise finite difference of $F_{G,x,y}$ for the group $G$. In Pythia-410M the groups had 32 components (two layers of heads, or all 24 MLP blocks), and in Pythia-1.4B four. To obtain exact single-component scores, we regroup by layer. A layer's 16 heads and its MLP block read the same layer input under Pythia's parallel residual, so patching one of them does not change the others, and each score is the effect of patching that component alone (no site is an ancestor of another, as in Section~\ref{sec:theory}). We recompute these scores on the same calibration items and evaluate set-patching recovery of their top-$k$ components on the same test items. Recomputing the reported rankings' recovery with this code reproduces the reported recovery values to within $0.006$ nats.

\begin{table}[h]
\centering
\caption{\textbf{Exact single-component scores versus the reported single-patch ranking.} Columns give the Spearman correlation between the two rankings, the number of MLP blocks among the top five of each, and the fraction of the dense NLL lift recovered on the test set by the top-$k$ components of the exact (reported) ranking. Pythia-410M entries are means over 3 seeds (Spearman range in parentheses), and Pythia-1.4B is one seed.}
\label{tab:exact}
\small
\setlength{\tabcolsep}{4pt}
\begin{tabular}{@{}llccccc@{}}
\toprule
Model & Domain & Spearman & MLPs in top 5 & $k=3$ & $k=5$ & $k=16$ \\
\midrule
410M & GSM8K & $0.82$ (0.81--0.83) & 4.7 (1.3) & 30\% (22\%) & 42\% (34\%) & 82\% (69\%) \\
410M & MMLU & $0.87$ (0.85--0.89) & 3.3 (3.3) & 88\% (74\%) & 97\% (93\%) & 99\% (98\%) \\
410M & PubMedQA & $0.92$ (0.90--0.95) & 5.0 (5.0) & 87\% (92\%) & 98\% (103\%) & 103\% (102\%) \\
1.4B & PubMedQA & $0.96$ & 5 (5) & 79\% (76\%) & 88\% (94\%) & 107\% (107\%) \\
\bottomrule
\end{tabular}
\end{table}

The two rankings agree closely in correlation but differ in their top few entries, mostly in how they order MLP blocks. The conclusion that a few components recover most of the answer-sequence NLL lift holds for both. With exact scores, five components recover 97\% on MMLU, 98\% on PubMedQA, and 88\% at 1.4B, and GSM8K remains the domain that needs many components. Table~\ref{tab:e2-aurc} and Figure~\ref{fig:e2-recovery} use the reported ranking.

\section{Ranking stability}
\label{app:stability}

A localization is more useful when it does not depend on incidental choices. This appendix varies the calibration sample, the prompt template, the scoring metric, and the training seed.

\paragraph{Calibration sample.} In the Pythia-1.4B run, we compare rankings from the first and second halves of the 256 shuffled calibration items. With model, specialist, and prompt fixed, they have full-rank Spearman correlation $0.978$, and top-$k$ Jaccard is $1.00$, $0.67$, $1.00$, $1.00$, and $0.88$ at $k=3$, $5$, $8$, $16$, and $32$.

\paragraph{Prompt template.} For the seed-0 Pythia-410M specialist of each domain, we recompute exact single-component scores (Appendix~\ref{app:exact}) on the same 500 calibration items under the original template and under an alternate template that rewords the instruction, renames and reorders the fields, and keeps the ``Answer:'' cue and the target tokens. The alternate templates are: GSM8K, ``Work out the answer to this grade-school math word problem.\textbackslash nProblem: \{question\}\textbackslash nAnswer:''; PubMedQA, ``Abstract: \{context\}\textbackslash nQuestion: \{question\}\textbackslash nUsing the abstract, reply to the question with yes, no, or maybe.\textbackslash nAnswer:''; MMLU, ``The following is a multiple-choice question about \{subject\}.\textbackslash n\{question\}\textbackslash nOptions:\textbackslash nA) \ldots\textbackslash nReply with the letter of the correct option.\textbackslash nAnswer:''. Table~\ref{tab:prompt} reports the agreement. The specialist's calibration NLL lift barely changes on GSM8K ($1.34$ nats under both templates) but falls on MMLU ($3.89$ to $3.48$) and PubMedQA ($9.52$ to $6.87$), consistent with fine-tuning on a single template.

\begin{table}[h]
\centering
\caption{\textbf{Agreement between exact single-component rankings under two prompt templates.} Pythia-410M, seed 0, 500 calibration items.}
\label{tab:prompt}
\small
\begin{tabular}{@{}lccccc@{}}
\toprule
Domain & Spearman & Top-3 Jaccard & Top-5 Jaccard & Top-8 Jaccard & Top-16 Jaccard \\
\midrule
GSM8K & $0.99$ & $1.00$ & $0.67$ & $0.78$ & $1.00$ \\
MMLU & $0.82$ & $1.00$ & $0.67$ & $1.00$ & $0.78$ \\
PubMedQA & $0.71$ & $0.50$ & $0.43$ & $0.60$ & $0.68$ \\
\bottomrule
\end{tabular}
\end{table}

\paragraph{Accuracy versus NLL scores.} In the same forward passes, we score each component by the change in option-likelihood accuracy on the calibration items (the fraction whose highest-probability allowed label at the answer position is the gold label). On MMLU the accuracy-based and NLL-based rankings are unrelated (Spearman $-0.07$, $-0.17$, and $0.00$ for seeds 0, 1, and 2, with top-5 Jaccard at most $0.11$), and 29--46\% of components leave calibration accuracy unchanged. On PubMedQA they are moderately related (Spearman $0.66$--$0.73$ across seeds and $0.55$ at 1.4B, with top-5 Jaccard $0.11$--$0.43$). The PubMedQA calibration items come from PQA-A, which is 92--93\% yes, and the specialists' calibration accuracy (0.91--0.95) mostly reflects predicting yes, so an accuracy gain there measures movement toward yes. The NLL and mean target log-probability scores are equivalent, and accuracy is a separate endpoint.

\paragraph{Across training seeds.} Rankings from specialists trained with different seeds, on different calibration draws, have mean pairwise full-rank Spearman correlation $0.96$ on GSM8K, $0.62$ on PubMedQA, and $0.44$ on MMLU, and mean top-8 Jaccard $0.78$, $0.78$, and $0.21$. These comparisons change the specialist and the calibration sample together.

\section{E3 extras: selection overlap, tiny-budget comparison, head-level adapters}
\label{app:e3-extras}

This appendix adds three comparisons to Section~\ref{sec:e3}. Table~\ref{tab:e3-jaccard} reports how much each method's adapted blocks, read from the saved adapters, overlap with the patching-localized blocks at the same (domain, seed, $k/N$). The overlaps are close to the expected overlap of two independent uniformly random block sets of the same size.

\begin{table}[h]
\centering
\caption{\textbf{Median Jaccard overlap of selected blocks with patching-localized}, pooled over domains and seeds (n${=}9$ per cell). The last row is the expected overlap of two independent random sets of $k$ out of 24 blocks.}
\label{tab:e3-jaccard}
\small
\begin{tabular}{lccc}
\toprule
Method & $k/N{=}0.25$ & $0.50$ & $0.75$ \\
\midrule
Random-localized   & 0.20 & 0.33 & 0.57 \\
Dense-update-norm  & 0.09 & 0.26 & 0.64 \\
Domain-probe       & 0.09 & 0.33 & 0.64 \\
Full LoRA          & 0.25 & 0.50 & 0.75 \\
Random sets (expected) & 0.15 & 0.34 & 0.60 \\
\bottomrule
\end{tabular}
\end{table}

Block choice could matter more when the budget is very small, so the tiny-budget comparison (PubMedQA and MMLU, $k\in\{1,2,3\}$ blocks, $r\in\{4,8\}$, 3 seeds, 10 random draws per cell) compares patching-localized against random-localized (median over draws) and dense-update-norm. The patching-minus-random NLL gap (positive favors random) is $+0.002$ $[-0.002, 0.006]$ on MMLU and $+0.053$ $[-0.008, 0.114]$ on PubMedQA, with both 95\% CIs including zero.

\paragraph{Head-level versus block-level adapters.}
E2 ranks individual heads while LoRA adapts whole blocks, so we also train an adapter restricted to individual heads. At the largest E3 budget on PubMedQA ($k/N{=}0.75$, $r{=}16$, 18 blocks), we compare the block-level adapter (attention and MLP projections of the 18 selected blocks, $4{,}718{,}592$ parameters) with a head-level adapter that applies rank-60 LoRA only to the query, key, value, and output projections of the highest-ranked head in each selected block ($4{,}700{,}160$ parameters). Test NLL is $1.010 \pm 0.107$ for the head-level adapter and $1.287 \pm 0.286$ for the block-level adapter (mean $\pm$ s.e.\ over 3 seeds). Paired by seed, the differences are $-0.61$, $0.00$, and $-0.21$ nats. Both adapters predict yes on every test item, so their option-likelihood accuracy equals the majority-label rate. Both were trained on the 256 E3 training examples.

\section{Reproducibility and asset notes}
\label{app:repro-assets}

The experiments use the public assets Pythia-410M and Pythia-1.4B (Apache License 2.0), GSM8K (MIT License), MMLU (MIT License), and PubMedQA (MIT License), loaded through the Hugging Face \texttt{transformers} and \texttt{datasets} libraries. Code, derived result tables, and the scripts that regenerate every table and figure are released at \url{https://github.com/AsiaeeLab/specialist-mediators}. All language-model experiments ran on single NVIDIA RTX 4500 Ada GPUs (24 GB). Single-patch scoring over 408 components and 500 calibration examples takes 13--27 minutes per domain and seed at 410M with float16 and a peak of 14 GB. Training one E3 adapter takes under one minute, and the full E3 grid of 405 adapters and its NLL re-evaluation take a few GPU-hours.

\end{document}

%% file: related_work_appendix.tex
This appendix surveys work further from our setting, including task- and domain-specific subnetworks, structured pruning, learned sharing and modularity, parameter-efficient adaptation, and causal abstraction. Each of these either constructs a compact or modular model or evaluates a single model, whereas our object is the difference between a base model and its dense specialist.

\paragraph{Task-specific subnetworks and parameter isolation.}
A direct route to specialization is to store multiple subnetworks inside one model. PackNet, Piggyback, and SupSup allocate task-specific masks or parameters via iterative pruning or learned binary masks, motivated by continual learning and parameter isolation \citep{mallya2018packnet, mallya2018piggyback, wortsman2020supsup}. The lottery ticket hypothesis and hidden-subnetwork results show that strong task-specific subnetworks exist even in randomly initialized or untrained weights \citep{frankle2019lottery, ramanujan2020hidden}. Domain-specific subnetwork extraction has been pursued directly in NMT---DoSS uses pruning-derived masks to assign a sub-network to each domain \citep{hendy2022doss}, and Prune-Tune learns tiny domain-specific sparse structures during fine-tuning \citep{liang2021prunetune}---and at LLM scale by D-Pruner \citep{zhang2024dpruner} and ATP, which couples tuning with structural pruning in a single stage \citep{lu2024atp}. These methods demonstrate that specialization can be realized by masks or pruning. They build the specialized subnetwork, whereas we ask which components of an already trained dense specialist carry its lift over the base.

\paragraph{Structured pruning with replacement.}
Classical pruning methods rank parameters by second-order sensitivity \citep{lecun1989obd, hassibi1992obs}. Modern compression targets \emph{structured} units (filters, channels, attention heads, blocks) because wall-clock gains require dense kernels after pruning rather than irregular sparsity \citep{han2016deepcompression, li2017pruning, liu2017network, he2017channel, louizos2018l0}. Related criteria include Taylor saliency, head importance, and activation statistics \citep{molchanov2019importance, michel2019heads, voita2019analyzing}. Mean-replacement pruning, neuron merging, and variance-based pruning are directly relevant because they replace deleted units by folded constants or nearby units rather than zeroing them \citep{evci2018mean, kim2020neuron, berisha2025vbp}. Causal mechanism reduction treats these as special cases of a second-order replacement-risk objective \citep{asiaee2026efficient}. For LLMs, one-shot Hessian-aware and magnitude-aware methods (OBC, SparseGPT, Wanda) extend post-training pruning to scale \citep{frantar2022obc, frantar2023sparsegpt, sun2024wanda}. LLM-Pruner discovers inter-dependency groups \citep{ma2023llmpruner} and ShortGPT removes whole transformer layers \citep{men2024shortgpt}. Our single-patch score borrows the interventional scoring idea but changes the target, measuring how much of the specialist's lift a component carries rather than how much the model's own loss depends on it.

\paragraph{Learned sharing, modularity, and experts.}
Multi-task and modular systems learn how to share representations or experts during training. Hard- and soft-sharing architectures such as cross-stitch networks \citep{misra2016crossstitch}, sluice networks for latent multi-task architecture learning \citep{ruder2019sluice}, and multi-gate mixtures of experts \citep{ma2018mmoe} learn task-specific mixing weights or gates over a shared backbone. Mixture-of-experts architectures learn sparse routing over explicitly parameterized experts \citep{shazeer2017moe, fedus2022switch, lu2024experts, liu2024eep}, and the modular deep learning paradigm organizes computation into autonomous parameter-efficient modules with conditional routing \citep{pfeiffer2024modular}. Whether trained networks themselves develop functionally modular substructure has been studied empirically. Differentiable weight masks reveal that standard architectures often fail to develop clean task-modules \citep{csordas2021modular}, and resource constraints and information-flow timing shape specialization dynamics \citep{bena2025dynamics}. We ask the complementary post-hoc question of which components mediate the difference between an already trained base and dense specialist, without training a new modular architecture.

\paragraph{Parameter-efficient adaptation and model merging.}
Adapters, low-rank updates, and localized interventions specialize on top of a frozen backbone \citep{houlsby2019adapters, pfeiffer2021adapterfusion, hu2021lora, wang2022adamix, liu2024dora}. Diff pruning enforces sparsity on a task-specific parameter diff via a differentiable $L_0$ penalty \citep{guo2021diffpruning}, and LoFiT first identifies a sparse set of attention heads and then fine-tunes only small offsets there \citep{yin2024lofit}. Model merging combines specialists in parameter space via task vectors, sign-aware merging, sparse masking, and subspace-aware merging \citep{ilharco2023task, yadav2023ties, yu2023dare, davari2024breadcrumbs, iurada2025talos, cheng2025wudi, skorobogat2025subspace}. See \citet{yang2024modelmerging} for a survey. These approaches operate primarily in parameter-delta or weight space, whereas we work with activations of individual components and score them by intervention effects on the specialist's lift.

\paragraph{Causal abstraction and mechanistic interventions.}
Causal abstraction studies when a high-level structural causal model is a faithful simplification of a lower-level one under interventions, with exact and approximate variants and soft-intervention extensions \citep{rubenstein2017causal, beckers2019abstracting, beckers2020approximate, massidda2023soft, geiger2021causal, geiger2025theoretical}. For neural networks, the central interventional primitive is the \emph{interchange intervention}, underlying interchange intervention accuracy, distributed alignment search, and causal similarity methods \citep{geiger2024finding, grant2025model}, with closely related localization tools such as causal tracing, path patching, and causal scrubbing \citep{meng2022locating, goldowskydill2023path, chan2022scrubbing}. Recent work shows that localization results are sensitive to patching metrics and corruption choices \citep{zhang2024bestpractices, conmy2023acdc, shi2024circuit}. We use this intervention semantics to localize a model difference rather than to certify a human-readable circuit.

\paragraph{Evaluation and uncertainty.}
Interventional claims are easiest to overstate when the search process is adaptive, since circuit metrics can move substantially with the ablation method or discrepancy function \citep{zhang2024bestpractices, shi2024circuit}. Certified interventional fidelity addresses this with confidence intervals and anytime-valid confidence sequences for interventional estimands \citep{asiaee2026certified}. This motivates treating interventional evaluation as part of the procedure rather than an afterthought, which is why we report the patching ceiling, absolute recovery, and ranking stability across calibration samples, prompt templates, and scoring metrics alongside normalized recovery curves.